\documentclass{article}

\usepackage[preprint,nonatbib]{neurips_2024}

\usepackage[utf8]{inputenc} 
\usepackage[T1]{fontenc}    
\usepackage{hyperref}       
\usepackage{url}            
\usepackage{booktabs}       
\usepackage{amsfonts}       
\usepackage{nicefrac}       
\usepackage{microtype}      
\usepackage{xcolor}         

\usepackage{booktabs} 
\usepackage{tabularx} 
\usepackage{array}
\usepackage{tcolorbox}
\tcbuselibrary{most}
\usepackage{graphicx}  
\usepackage{caption}   
\usepackage{subcaption} 
\usepackage{multirow}
\usepackage{comment}

\title{LLM-RadJudge: Achieving Radiologist-Level Evaluation for X-Ray Report Generation}

\author{%
  Zilong Wang, Xufang Luo, Xinyang Jiang, Dongsheng Li, Lili Qiu\\
  Microsoft Research Asia\\
  \texttt{\{wangzilong,xufluo,xinyangjiang,dongsheng.li,liliqiu\}@microsoft.com} \\
}

\begin{document}

\maketitle

\begin{abstract}
Evaluating generated radiology reports is crucial for the development of radiology AI, but existing metrics fail to reflect the task's clinical requirements. This study proposes a novel evaluation framework using large language models (LLMs) to compare radiology reports for assessment. We compare the performance of various LLMs and demonstrate that, when using GPT-4, our proposed metric achieves evaluation consistency close to that of radiologists. Furthermore, to reduce costs and improve accessibility, making this method practical, we construct a dataset using LLM evaluation results and perform knowledge distillation to train a smaller model. The distilled model achieves evaluation capabilities comparable to GPT-4. Our framework and distilled model offer an accessible and efficient evaluation method for radiology report generation, facilitating the development of more clinically relevant models. The model will be open-sourced and accessible at [placeholder].
\end{abstract}

\section{Introduction}
Radiology is a highly specialized field with a significant demand for accurate and timely diagnoses, where the increasing need is underscored by a shortage of medical professionals. Recent advancements in artificial intelligence have shown promising results in optimizing radiology image analysis, which is expected to improve healthcare accessibility and deliver precise diagnosis~\cite{Harris:2019,ScottJ.Adams:2021,Alvarado:2022}. Large language models (LLMs) and vision-language models have made it possible to further generate radiology reports from image in a nature language form~\cite{driess2023palm,wang2023chatcad,hyland2023maira}. To effectively advance this research, Peter Drucker's famous saying, 'What gets measured gets managed,' highlights the importance of evaluation metrics in the task of report generation. As training scales increase, it becomes impractical to involve manual evaluation from radiologist for every training strategy and model selection details, making automated evaluation methods crucial.

Generated report evaluation is a task where the goal is to evaluate the quality of generated radiology reports by comparing them against a reference ground truth report. This involves a careful examination to determine their clinical relevance and accuracy.
Current evaluation metrics for radiology reports can be  categorized into two main types: language metrics such as BLEU~\cite{papineni2002bleu}, ROUGUE~\cite{lin2004rouge} and METOER~\cite{banerjee2005meteor} scores, focusing on aspects like n-gram overlap, sequence alignment, and synonym matching to assess adequacy. They are widely used in numerous tasks as well as report generation by comparing their similarity to reference ground truth report. Also, similarity comparison by models like BERTScore is also used~\cite{zhang2019bertscore}. 
The second type is clinical metrics, like CheXpert F1~\cite{smit2020combining} and RadGraph F1~\cite{jain2021radgraph}. They focus more on clinical events described in the radiology report, such as pathological entities, their location and severity based on some pre-defined types of conditions.

Although widely adopted in existing methods, both types of metrics have significant limitations when evaluating radiology reports that require consideration of clinical significance. language metrics are primarily confined to grammatical and lexical similarities, which may not adequately represent the accuracy of reports in clinical diagnostic scenarios. Clinical metrics using a limited set of predefined categories struggle to encompass the complex spectrum of clinical possibilities in reports. Moreover, these existing clinical metrics struggle to properly assess the partial correctness on inclusive relationships (left upper lobe and left lung), near synonyms (nodule and opacity) and equivalent expressions (cardiomegaly and enlarged heart). Even composed approaches did not address the issues with a satisfactory results~\cite{yu2023evaluating,calamida2023radiology}. Establishing a standardized terminology system including hierarchical relationships and correlations that uniformly represents all diagnoses is too complex to achieve. Large language models, with their expansive knowledge base, offer a more nuanced and flexible understanding of text, allowing for the discernment of subtle distinctions. This inspires us to explore new methods in our approach.

Large language models like GPT-4 have demonstrated formidable capabilities in various domains~\cite{achiam2023gpt}, particularly in the analysis and comparison of texts. Research has also shown that these models have substantial potential in the medical field~\cite{garikipati2024openmedlm}. Motivated by these findings, we explored the utility of large language models for the assessment of radio logical reports and reached a near radiologist-level performance. Moreover, these models offer detailed description and classification of errors, enhancing the interpretability of assessments and enabling deeper analysis. 

However, models like GPT-4 have limitations in response time and costs, making them unsuitable for large-scale usage, especially during model development and testing, which require frequent testing. To address this, we propose developing a efficient language model with smaller size achieves similar performance to GPT-4 but with significantly faster response, larger throughout and lower costs, making it an accessible metric. The enhanced performance, as evidenced by a stronger Kendall's tau correlation with human expert annotations, may stem from continuous pre-training on biomedical corpus and the strategic curation of training data quality. 

Our main contributions include:
\begin{itemize}
    \item We propose an effective methodology, LLM-RadJudge, using LLMs to evaluate radiology reports. 
    \item We introduce a smaller model that matches the efficacy of GPT-4 with accessibility
    \item We have conducted a series of experiments to validate the performance of both the large language model and the smaller model, comparing them with manual assessments by radiologists.
\end{itemize}

\section{Related Work}

\subsection{Evaluation Metrics for Radiology Reports}

There are lots of metrics proposed for evaluating radiology reports. Same as us, most of them focus on X-ray reports due to the availability of the large-scale public dataset~\cite{johnson2019mimic}. These metrics can be generally divided into two groups.
\paragraph{Language metrics}
The first group contains some metrics for general nature language generation evaluation, including but not limiting to BLEU~\cite{papineni2002bleu}, ROUGUE~\cite{lin2004rouge} and METOER~\cite{banerjee2005meteor} scores.
Or some other scores calculated using embedding generated by pre-trained models, like BERTScore~\cite{zhang2019bertscore}.
These are included in most research works on the report generation for measuring the similarity between the generated report and ground-truth report. But as they have no special attention on clinical events described in the radiology reports, leading to little clinical significance.
\paragraph{Clinical metrics}
Unlike general language metrics, the second group emphasizes clinical descriptions in the radiology report, and thus is more useful and meaningful for real-world applications. The clinical descriptions reflect all clinical events in the corresponding medical image, including but not limiting to the pathological entity, location and severity. One of the most common metric in current works is CheXpert, in which each of the 14 CheXpert pathological entities should be extracted and labelled with `present/absent/uncertain'. Subsequently, the accuracy of these classifications is assessed. This is done with some labelers, like CheXbert~\cite{smit2020combining}. Cosine similarity from labeler's embedding also acts as a metric. Another metric is RadGraph~\cite{jain2021radgraph,khanna2023radgraph2}, which extracts clinical entities and relations from the report using the RadGraph model. These extraction-based methods are limited by the pre-defined entity set or the rigid matching rule, leading to the coverage issue or incapability of handling widely existed vague cases in the report~\cite{jain2021radgraph}. Some hybrid methods try to combine existing metrics, like RadCliQ~\cite{yu2023evaluating} using a linear function, and RadEval~\cite{calamida2023radiology} using a trained model. But they still cannot overcome the shortcomings of extraction-based methods on evaluating clinical descriptions. Different from these work, our method leverages large language models for evaluation, which largely enhances flexibility and can reach the radiologist-level.

\subsection{Large Language Model for Evaluation}

Using large language models as an automatic evaluation metric is explored in some previous work, such as G-Eval~\cite{liu2023g} and LLM Evaluation~\cite{chiang-lee-2023-large}. The recent investigation~\cite{zheng2024judging} shows that such LLM-as-a-Judge methods perform differently on different tasks. Previous works focus on general language generation tasks, and to best of our knowledge, none of these works targets radiology report evaluation, which requires specific designs to make the results clinically meaningful.

\section{Report Evaluation Using Large Language Models (LLMs)}

\subsection{Motivating Examples} 

\begin{figure}

\begin{minipage}{\textwidth}
\begin{subfigure}[b]{\textwidth}
\tcbset{colback=white,arc=0mm,width=(\linewidth-4pt)/2,
equal height group=AT1,
before=,after=\hfill,fonttitle=\bfseries}
\noindent
\begin{tcolorbox}[title=GT: A normal report,colframe=green!75!black]
FINDINGS: Frontal and lateral views of the chest were obtained. No focal consolidation, pleural effusion, or evidence of pneumothorax is seen. The cardiac silhouette is top normal. Mediastinal and hilar contours are unremarkable. No displaced fracture is seen. There is no evidence of free air beneath the diaphragms.
\end{tcolorbox}
\begin{tcolorbox}[title=GR1: Equivalence report expressed differently]
FINDINGS: AP and lateral CXR. No indications of focal consolidation or pleural effusion is seen. There is no evidence of pneumothorax. The cardiac outline is within the upper limits of normal. The mediastinal and hilar contours do not show any abnormalities. No evidence of a fracture is present. Additionally, there is an absence of free air beneath the diaphragms areas.
\end{tcolorbox}
\tcbset{equal height group=AT2}
\begin{tcolorbox}[title=GR2: Lateral views absent\text{,} numerous additional findings]
FINDINGS: Frontal views of the chest were obtained. Focal consolidation, pleural effusion, and evidence of pneumothorax are seen. The cardiac silhouette is enlarged. Mediastinal and hilar contours are unremarkable. Rib fracture is seen. There is no evidence of free air beneath the diaphragms.
\end{tcolorbox}
\begin{tcolorbox}[title=GR3: Different cardiac silhouette\text{,} free air]
FINDINGS: Frontal and lateral views of the chest were obtained. No consolidation, no pleural effusion, no pneumothorax. The cardiac silhouette is bottom normal. Mediastinal and hilar contours are unremarkable. No fracture. There is some evidence of free air beneath the diaphragms.
\end{tcolorbox}
\caption{An abnormal reference report (GT) and three corresponding comparative reports.}
\end{subfigure}

\vspace{1em}

\begin{subfigure}[b]{\textwidth}
\tcbset{colback=white,arc=0mm,width=(\linewidth-4pt)/2,
equal height group=AT3,
before=,after=\hfill,fonttitle=\bfseries}
\noindent
\begin{tcolorbox}[title=GT: A case with enlarged heart\text{,} increased multiple opacity and old rib fractures,colframe=green!75!black]
Heart size is enlarged. Mediastinum is stable. Multifocal opacities are present, overall similar to previous study but potentially minimally improved. No appreciable pneumothorax. Old rib fractures, unchanged.
\end{tcolorbox}
\begin{tcolorbox}[title=GR1: Equivalence of GT in different writing.]
Cardiomegaly is found. Mediastinal silhouette remains unchanged. There are multiple areas of increased opacity within the lungs, which appear largely consistent with the prior examination, with a slight possibility of marginal improvement. No significant evidence of pneumothorax. Prior rib fractures are present with no interval change.
\end{tcolorbox}
\tcbset{equal height group=AT4}
\begin{tcolorbox}[title=GR2: Single decreased opacity]
Mediastinum and heart size is enlarged. Mediastinum is stable. Single opacities are present, overall similar to previous study but potentially minimally decreased.No appreciable pneumothorax. Old rib fractures, unchanged.
\end{tcolorbox}
\begin{tcolorbox}[title=GR3: Decreased opacities and changed rib fractures]
Heart size is enlarged and mediastinum is large. Opacities are present in basiliar side,  overall similar to previous study but potentially minimally decreased. No pneumothorax. Old rib fractures, changed.
\end{tcolorbox}
\caption{A reference report(GT) and three corresponding comparative reports.}
\end{subfigure}

\end{minipage}
  \caption{Two group of cases of reference reports and generated reports}
  \label{fig:two_cases}
\end{figure}

\begin{table}
\begin{tabular}{l*{6}{c}}
\toprule
Group &  Candidate  & BLEU-2$\uparrow$ & BERTscore$\uparrow$ & CheXbert$\uparrow$ & RadgraphF1$\uparrow$ & RadCliQ$\downarrow$ \\
\midrule
\multirow{3}{*}{(a):normal} & GR1  & 0.472$\ast$                  & 0.728$\ast$                & 0.787                  & 0.600$\ast$                           & 1.52$\ast$                                      \\
                   & GR2  & 0.778                  & 0.886                & 0.305$\ast$                 & 0.671                         & 0.829                                     \\
                   & GR3  & 0.717                  & 0.903                & 0.965                  & 0.722                         & 0.854                                     \\
\midrule
\multirow{3}{*}{(b):abnormal} & GR1  & 0.114$\ast$                  & 0.486$\ast$                & 0.792$\ast$                  & 0.196$\ast$                           & 2.889$\ast$                                      \\
                   & GR2  & 0.850                  & 0.869                & 0.942                  & 0.602                         & 0.819                                     \\
                   & GR3  & 0.634                  & 0.794                & 0.858                  & 0.511                         & 1.375                                     \\
\bottomrule
\end{tabular}

\caption{Evaluation Scores on two group of cases: $\ast$ shows the worst score}
\label{tab:metrics_on_samples}
\end{table}

To illustrate the limitations of current evaluation metrics in assessing report generation, we present and analyze two group of cases where these metrics fail to correctly evaluate the generated reports. 

Consider two groups of generated reports in comparison with their ground truths, one depicting normal findings and the other showing abnormalities, as presented in Figure~\ref{fig:two_cases}. In these groups, the initial generated report (GR1) closely mirrors the clinical data from the ground truth (GT), conveying the same scenario with varied wording and representations. However, those interchangeable reports get almost the lowest scores on all evaluation metrics in Table~\ref{tab:metrics_on_samples}. On the other hand, minor variations in wording can result in significant changes in clinical meanings. For example, in group (a) GR2, only a few words were omitted, yet this omission led to the loss of meaning associated with the lateral view and the introduction of numerous additional findings.

\subsection{Architecting LLM-Based Methodology for Radiology Report Assessment}

Upon reviewing the aforementioned examples, it becomes clear that a profound understanding of the clinical ramifications is cardinal in crafting a diagnostic report, transcending mere lexical resemblances. Therefore, a meticulous comparative analysis of the clinical implications inherent in two diagnostic reports stands out as the cornerstone within the assessment framework. Within the peer review framework known as RADPEER, developed by the American College of Radiology (ACR), the scoring of comparisons is determined based on the nature and clinical relevance of the identified discrepancies\cite{abujudeh2014radpeer}. Additional evaluative studies have employed categorized error counts as a metric for radiologists' report assessment\cite{yu2023evaluating,calamida2023radiology}. We conceptualized the report evaluation task as one of categorized error detection and quantification. Specifically, in our experiments, we have utilized six predefined error categories as delineated in the ground truth labeling of the ReXVal dataset. These six error categories are as follows:
\begin{enumerate}
    \item False prediction of finding
    \item Omission of finding
    \item Incorrect location/position of finding
    \item Incorrect severity of finding
    \item Mention of comparison that is not present in the reference impression
    \item Omission of comparison describing a change from a previous study
\end{enumerate}
Furthermore, each category is subdivided into two subcategories: clinically significant and clinically insignificant errors, enabling a nuanced evaluation of the diagnostic reports.

Large language models (LLMs) have demonstrated remarkable proficiency in language annotation\cite{gilardi2023chatgpt,tornberg2023chatgpt}, and assessments of their performance reveal promising potential for applications within the medical domain\cite{garikipati2024openmedlm}.We hypothesize that Large Language Models (LLMs) are well-suited for report evaluation task. 

Notably, research in this domain has frequently omitted the detailed analytical steps that precede final outcomes, focusing predominantly on final quantitative results to economize on annotation time. However, it is critical for LLMs to process tasks "step-by-step", allowing enough space for reasoning. To optimize the functionality of LLMs in this context, our study introduces a prompt strategy contains two stages: The first stage involves a meticulous comparison of candidate reports against references to identify and list all errors. The second stage summarize these disparities to yield an aggregate total count as the score. This prompt design is informed by the principles of the Chain-of-Thought and Chain-of-Density\cite{wei2022chain,adams2023sparse}. This design not only steadies the inference process but also renders the LLMs' outputs in a more interpretable form. 

\begin{figure}[htbp]
    \centering
    \includegraphics[width=1\linewidth]{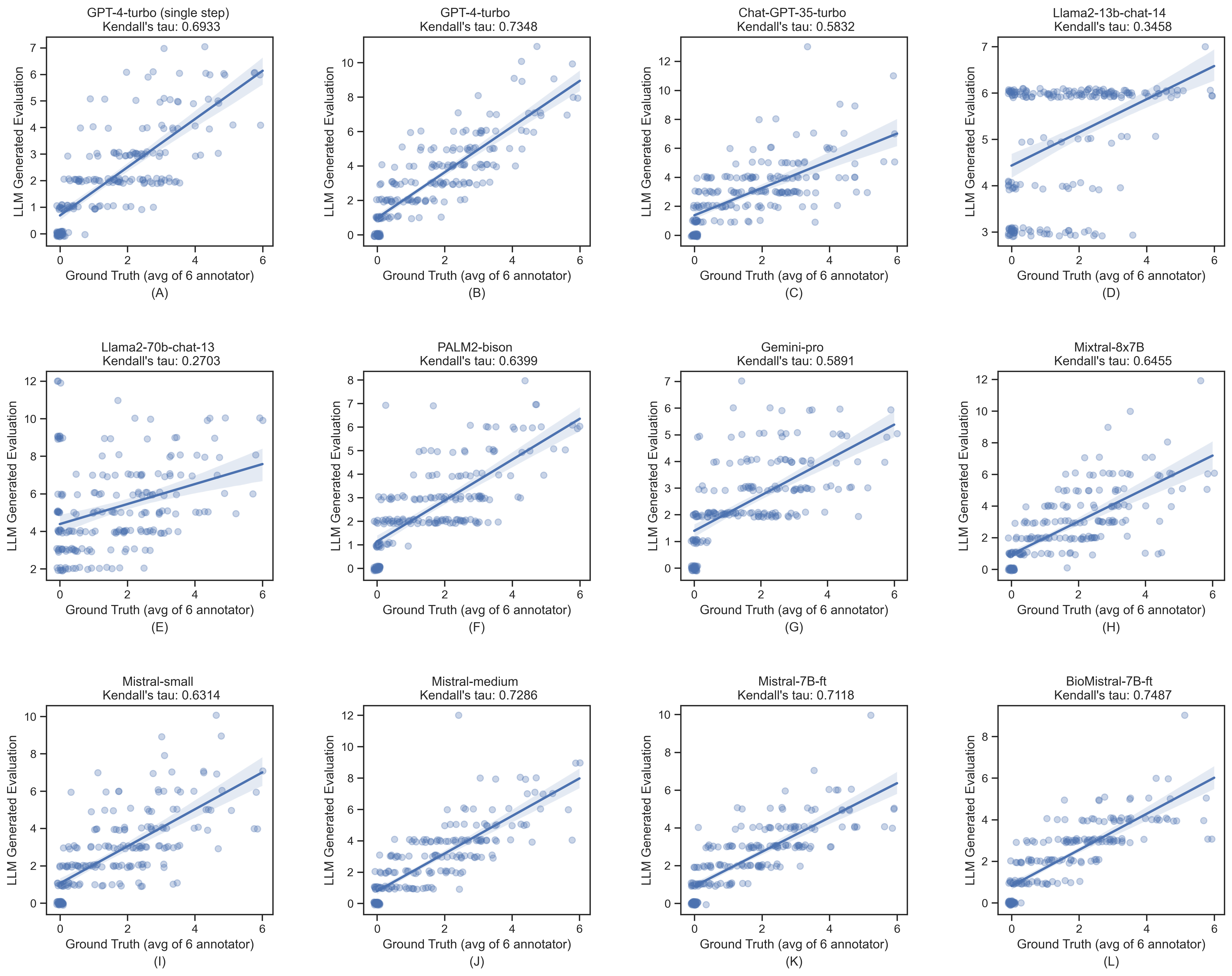}
    \caption{Accordance between LLMs and radiologists\\(A) is GPT-4 with single step prompt; (B)-(J) are different LLMs generated results; (K) is fine-tuned Mistral-7B; (L) is fine-tuned BioMistral-7B}
    \label{fig:llms}
\end{figure}

We evaluated both two-stages prompt and single-step prompt based on GPT-4. The results, as illustrated in Figure.\ref{fig:llms} (A) and (B), indicate an improvement associated with the two-stages prompting technique. Specifically, there was an increase in Kendall's tau correlation from 0.6933 to 0.7348, underscoring the efficacy of this approach.

\subsection{Benchmarking LLMs Alignments with Radiologist Expertise}

\begin{figure}[htbp]
    \centering
    \includegraphics[width=1\linewidth]{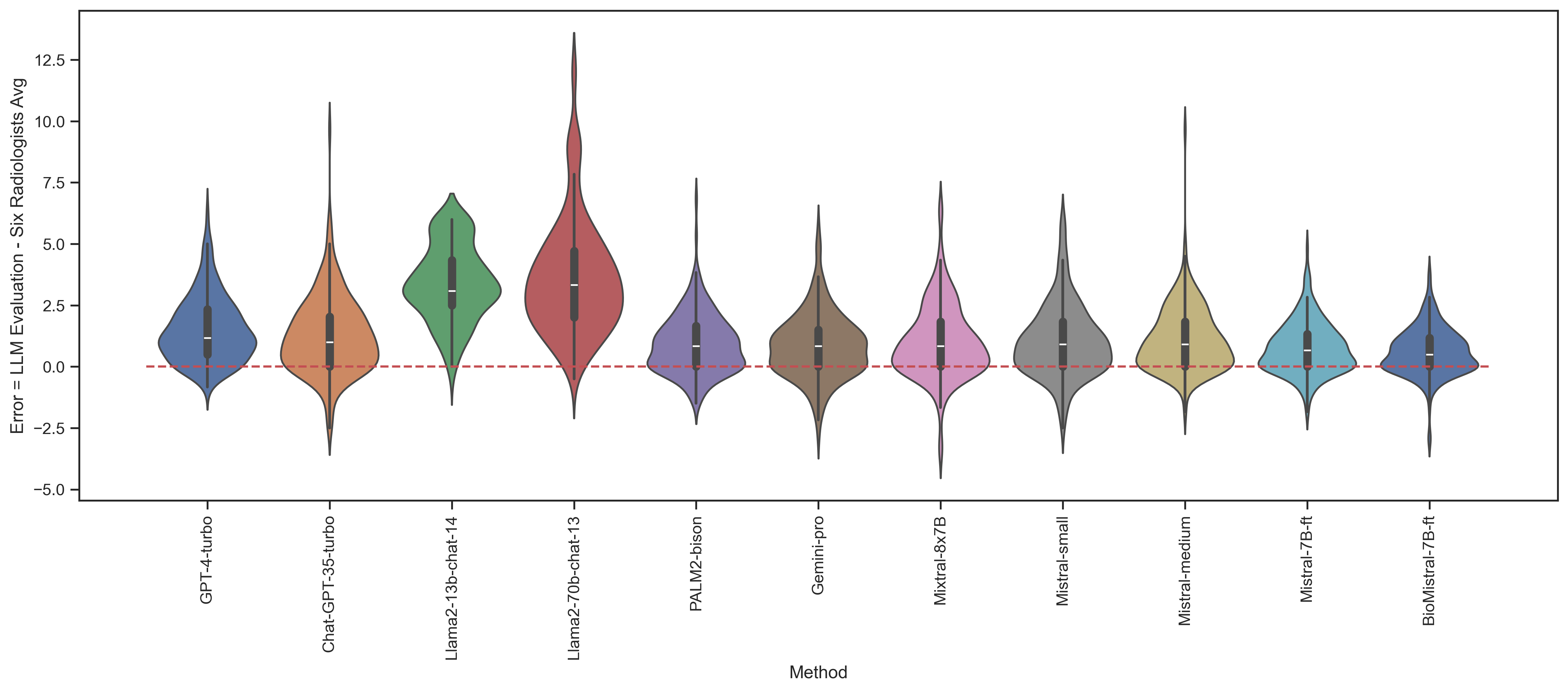}
    \caption{Error distribution of LLMs\\ Error = LLM evaluation - average counts of six radiologist}
    \label{fig:errors}
\end{figure}

Subsequently, we extended our comparative analysis to encapsulate a cohort of prevalent large language models (LLMs) to ascertain their efficacy in the report evaluation task. Besides GPT-4, this cohort encompassed an array of large language models such as GPT-3.5-turbo, PALM-2-bison, Gemini-pro, Llama2-13b-chat, Llama2-70b-chat, Mistral-small, Mixtral-8x7B, and Mistral-medium\cite{achiam2023gpt,anil2023palm,team2023gemini,touvron2023llama,jiang2023mistral,jiang2024mixtral}. 

We conducted a juxtaposition of the outputs produced by these LLMs against the annotations provided by radiologists from the ReXVal dataset and computed the Kendall's tau coefficient for each respective model, thereby assessing their concordance \cite{yu2023evaluating}. As delineated in Figure \ref{fig:llms}, panels (B) through (J), GPT-4-turbo emerged as the paragon, achieving the highest alignment with a Kendall's tau coefficient of 0.7348, closely tailed by Mistral-medium at 0.7286. The performance continuum then spans models such as Mixtral-8x7B, PALM2-bison, and Mistral-small, which exhibit Kendall's tau values ranging from 0.6455 to 0.6314. Models GPT-3.5-turbo and Gemini-pro hover around a Kendall's tau of approximately 0.58. In stark contrast, both Llama2 variants, Llama2-70b-chat and Llama2-13b-chat, demonstrate sub-optimal concordance. 

Despite the relatively strong performance of the Mistral series, issues with their output formats were identified, such as missing JSON keys or incorrect usage of quotation marks. We addressed these inconsistencies manually to ensure data integrity. 

Moreover, we charted the error distributions for various LLMs in Fig.~\ref{fig:errors}, which showed that these models tend to identify more differences between candidate and reference reports than manual annotations do, even with the models' varying levels of concordance with radiologists.

\subsection{Achieving Radiologist-Level Evaluation with LLM}

\begin{figure}[htbp]
    \centering
    \includegraphics[width=1\linewidth]{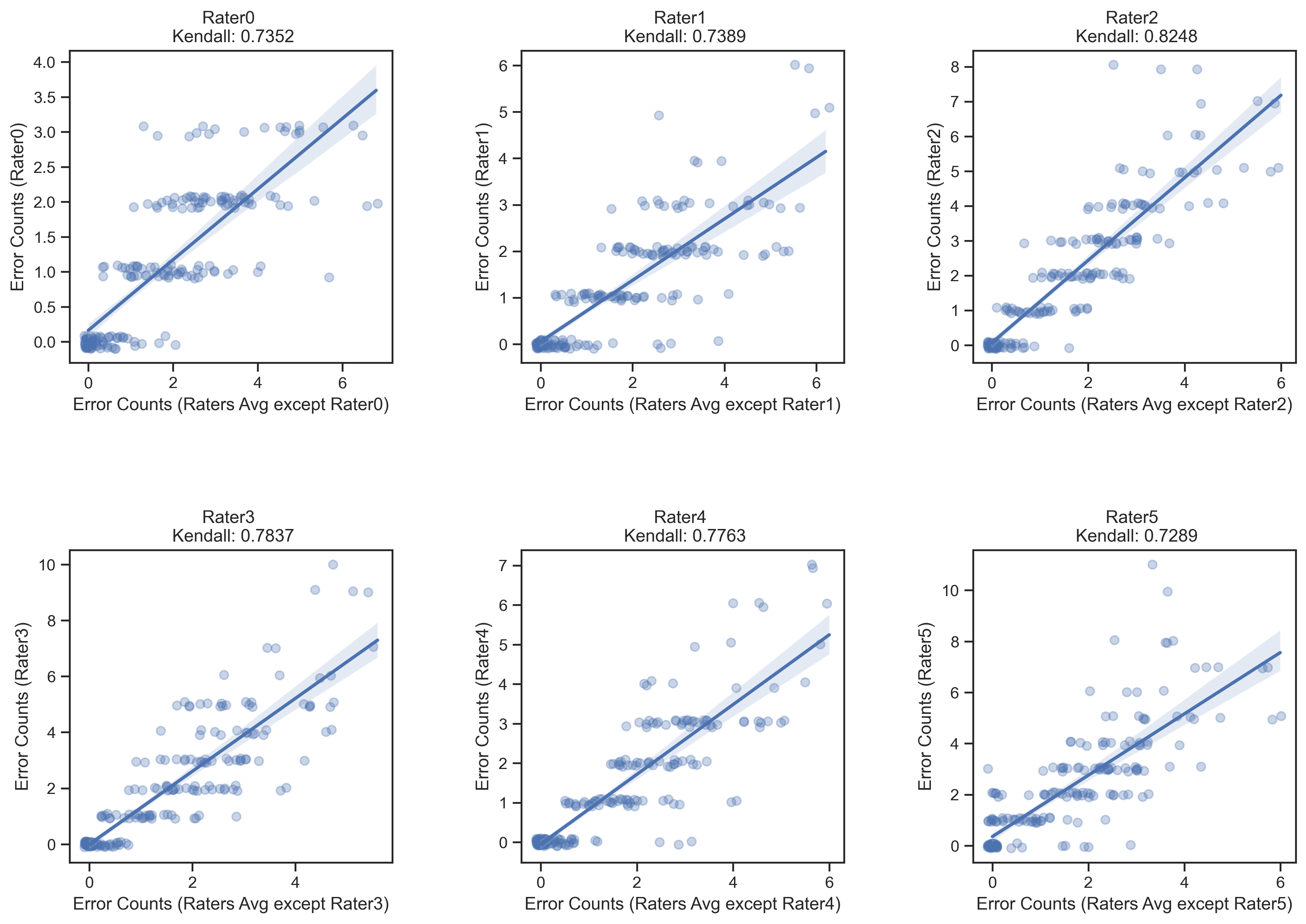}
    \caption{Accordance between radiologists}
    \label{fig:raters}
\end{figure}

GPT-4 demonstrated the highest performance among all LLMs, achieving a Kendall's tau correlation of 0.7348, surpassing various metrics such as BLEU, BERTScore, CheXbert vector similarity, RadGraph F1, and the newly proposed RadCliQ(Kendall's tau ranging from 0.414 to 0.615) on the test dataset . 

To further assess GPT-4's capabilities comparing with professional radiologists, we conducted an analysis of inter-observer consistency among the six radiologist annotators responsible for establishing the ground truth. We evaluated the concordance between each individual radiologist's assessments and the collective evaluations of the remaining five annotators. The resulting Kendall's tau values ranged from 0.7352 to 0.8248, indicating a high level of agreement among the radiologists. 

Remarkably, GPT-4's performance exceeded that of rater 5 and closely matched the evaluations of rater 0 and rater 1. These findings suggest that GPT-4's evaluation precision is nearly equivalent to that of human radiologists, demonstrating its potential to provide radiologist-level assessments.

However, it must be acknowledged that GPT-4 comes with its own set of limitations. The high cost and the slower response time significantly constrain the widespread application of this approach during the model training and optimization processes. 

\section{Developing an Efficient Model for Accessible Evaluation}

In response to the prohibitive expenses and time demands associated with GPT-4, previous researchers have attempted to distill its capabilities into smaller models for efficient usage while maintaining good performance\cite{zhou2023universalner,hu2024large}. Base on that insights, we aim to develop an efficient model that encapsulates the evaluative power of larger, high-performance models in a more compact form. Our goal is to make LLM-based report evaluation a practical metric that can aid report generation research by offering a affordable, fast, and accessible tool for evaluation.

\subsection{Curating Dataset for Model Training}

The dataset is critical for supervised fine-tuning of large language models (LLMs). We designed a dataset using a two-fold strategy. First, we selected reports and matched them with their top BLEU score counterparts to generate pairs of similar reports. Second, we randomly created diverse report pairs to enhance the generalizability and robustness of the comparison task. After sampling the outputs, we employed ChatGPT-35-turbo to swiftly screen out any records with apparent artifacts.

To construct pairs for comparison, we leveraged the MIMIC-CXR dataset, which contains 227,835 chest X-ray reports. We chose approximately 3,700 BLEU-paired reports and 7,500 randomly paired reports, and applied the previously described methods using GPT-4 to get evaluation results. Following quality control measures, the process yielded a refined dataset of 10,197 paired records and GPT-4 evaluation results, including the comparative analysis and final count data.

\subsection{Knowledge Distillation: Methodology and Experimental Setup}

We conducted supervised fine-tuning on the Mistral-7B-Instruct-v0.1 and BioMistral-7B models using Low-Rank Adaptation (LoRA) for parameter-efficient fine-tuning (PEFT) in the unsloth training framework. The training process was carried out on a single Nvidia A100 GPU machine with an effective batch size of 32 and a sequence length capped at 2048. We applied a learning rate of 1e-4 with cosine annealing and utilized the AdamW optimizer with 8-bit precision for 2 epochs. The finalized model was saved in bfloat16 format to enable high-throughput inference with serving tools.

\subsection{Evaluating the Performance of the Efficient Model} 

\begin{figure}[hbt]
    \centering
    \includegraphics[width=0.75\linewidth]{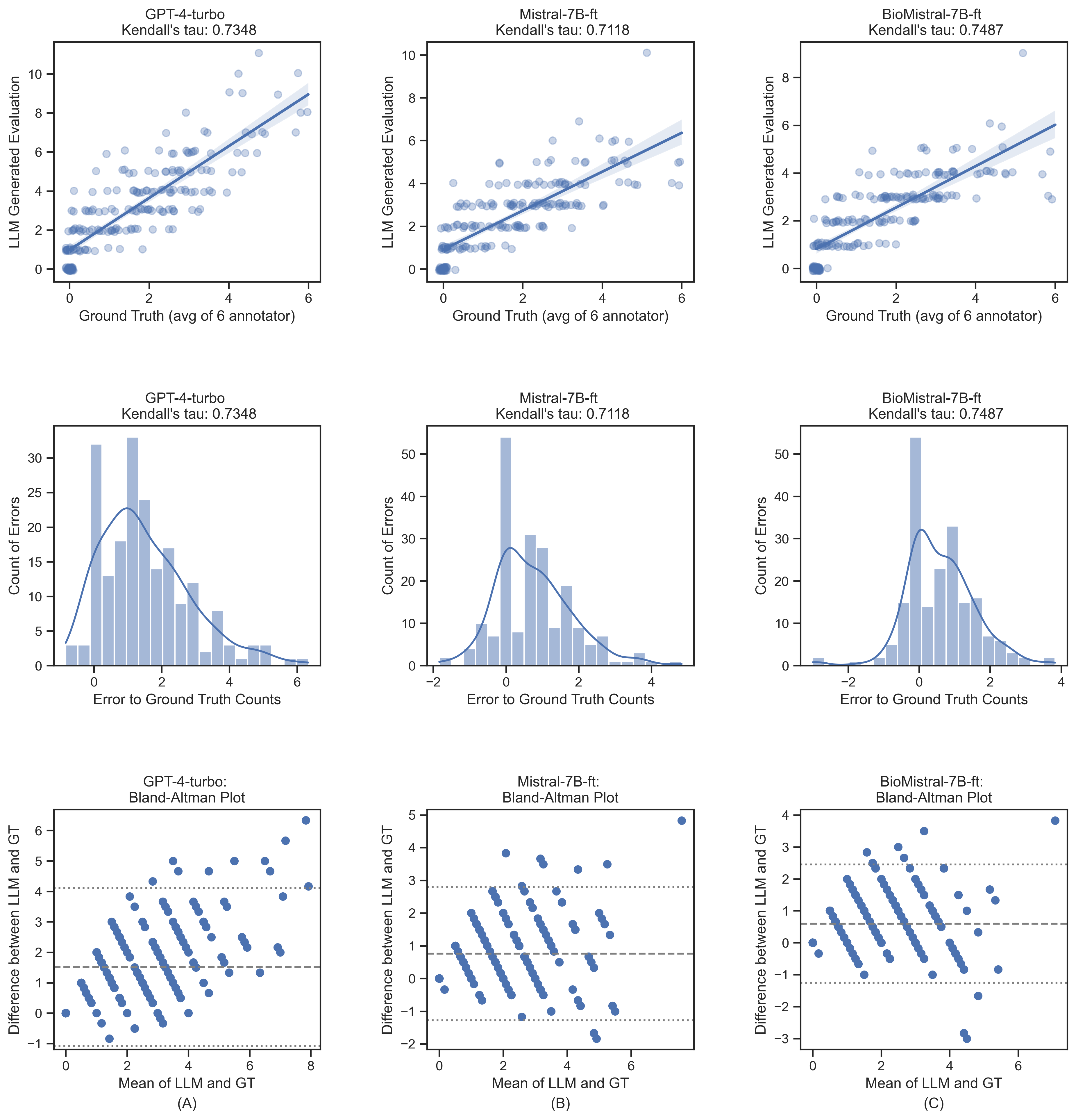}
    \caption{Comparison between GPT4 and finetuned models}
    \label{fig:tuned_comp}
\end{figure}

After distillation, panels (K) and (L) in Fig.~\ref{fig:llms} reveal that the Kendall's tau correlations for Mistral-7B and BioMistral-7B reached 0.7118 and 0.7487, respectively. The fine-tuned BioMistral-7B demonstrated marginally superior performance in alignment with radiologist assessments. Fig.~\ref{fig:tuned_comp} presents a comprehensive comparison of the characteristics of GPT-4, fine-tuned Mistral-7B, and BioMistral-7B in evaluation task, including scatter plots, error distributions, and Bland-Altman plots. Notably, BioMistral-7B exhibited not only a higher correlation but also a more normal error distribution when contrasted with GPT-4's left-skewed error distribution. Additionally, the Bland-Altman plot suggests a potential systematic increase in error as the magnitude of differences grows in GPT-4. This advantage of fine-tuned BioMistral-7B may be attributed to the target-focused continuous pretraining on biomedical datasets, which likely enhanced its domain-specific knowledge and ability to make more accurate predictions in the biomedical context.

\section{Conclusion}
This study presents a novel approach to evaluating radiology reports using large language models(LLMs) demonstrates performance comparable to that of radiologists. LLMs particularly GPT-4 showed a level of alignment with radiologists' assessments that surpasses existing metrics and approaches the nearly performance of individual radiologists.

Furthermore, to address the practical challenges associated with evaluating radiology reports using GPT-4, including high costs and slow response times, we have constructed a dataset derived from LLM evaluation results and performed distillation to fine-tune a smaller 7B model. The resulting distilled model, built upon BioMistral-7B, demonstrates evaluation capabilities that are comparable to, or even surpass, those of GPT-4, while offering significantly faster response times and lower computational costs. This makes our proposed evaluation method more feasible for widespread adoption in the development and assessment of radiology report generation models.

The proposed method has the potential to contribute to the field of report generation research. This tool is designed to reduce the reliance on radiologists' extensive involvement in the iterative processes of model optimization and selection, thereby streamlining the creation of models with enhanced clinical applicability. We plan to release the fine-tuned model to the community to facilitate further innovation in this sphere.

\bibliography{neurips_2024}

\begin{thebibliography}{10}

\bibitem{abujudeh2014radpeer}
H.~Abujudeh, R.~S. Pyatt~Jr, M.~A. Bruno, A.~L. Chetlen, D.~Buck, S.~K. Hobbs, C.~Roth, C.~Truwit, R.~Agarwal, S.~T. Kennedy, et~al.
\newblock Radpeer peer review: relevance, use, concerns, challenges, and direction forward.
\newblock {\em Journal of the American College of Radiology}, 11(9):899--904, 2014.

\bibitem{achiam2023gpt}
J.~Achiam, S.~Adler, S.~Agarwal, L.~Ahmad, I.~Akkaya, F.~L. Aleman, D.~Almeida, J.~Altenschmidt, S.~Altman, S.~Anadkat, et~al.
\newblock Gpt-4 technical report.
\newblock {\em arXiv preprint arXiv:2303.08774}, 2023.

\bibitem{adams2023sparse}
G.~Adams, A.~Fabbri, F.~Ladhak, E.~Lehman, and N.~Elhadad.
\newblock From sparse to dense: Gpt-4 summarization with chain of density prompting.
\newblock {\em arXiv preprint arXiv:2309.04269}, 2023.

\bibitem{ScottJ.Adams:2021}
S.~J. Adams, R.~D.~E. Henderson, X.~Yi, and P.~Babyn.
\newblock Artificial intelligence solutions for analysis of x-ray images.
\newblock {\em Canadian Association of Radiologists Journal}, 72(1):60--72, 2021.
\newblock PMID: 32757950.

\bibitem{Alvarado:2022}
R.~Alvarado.
\newblock Should we replace radiologists with deep learning? pigeons, error and trust in medical ai.
\newblock {\em Bioethics}, 36(2):121--133, 2022.

\bibitem{anil2023palm}
R.~Anil, A.~M. Dai, O.~Firat, M.~Johnson, D.~Lepikhin, A.~Passos, S.~Shakeri, E.~Taropa, P.~Bailey, Z.~Chen, et~al.
\newblock Palm 2 technical report.
\newblock {\em arXiv preprint arXiv:2305.10403}, 2023.

\bibitem{banerjee2005meteor}
S.~Banerjee and A.~Lavie.
\newblock Meteor: An automatic metric for mt evaluation with improved correlation with human judgments.
\newblock In {\em Proceedings of the acl workshop on intrinsic and extrinsic evaluation measures for machine translation and/or summarization}, pages 65--72, 2005.

\bibitem{calamida2023radiology}
A.~Calamida, F.~Nooralahzadeh, M.~Rohanian, K.~Fujimoto, M.~Nishio, and M.~Krauthammer.
\newblock Radiology-aware model-based evaluation metric for report generation.
\newblock {\em arXiv preprint arXiv:2311.16764}, 2023.

\bibitem{chiang-lee-2023-large}
C.-H. Chiang and H.-y. Lee.
\newblock Can large language models be an alternative to human evaluations?
\newblock In {\em Proceedings of the 61st Annual Meeting of the Association for Computational Linguistics (Volume 1: Long Papers)}, 2023.

\bibitem{driess2023palm}
D.~Driess, F.~Xia, M.~S. Sajjadi, C.~Lynch, A.~Chowdhery, B.~Ichter, A.~Wahid, J.~Tompson, Q.~Vuong, T.~Yu, et~al.
\newblock Palm-e: An embodied multimodal language model.
\newblock {\em arXiv preprint arXiv:2303.03378}, 2023.

\bibitem{garikipati2024openmedlm}
A.~Garikipati, J.~Maharjan, N.~P. Singh, L.~Cyrus, M.~Sharma, M.~Ciobanu, G.~Barnes, Q.~Mao, and R.~Das.
\newblock Openmedlm: Prompt engineering can out-perform fine-tuning in medical question-answering with open-source large language models.
\newblock In {\em AAAI 2024 Spring Symposium on Clinical Foundation Models}, 2024.

\bibitem{gilardi2023chatgpt}
F.~Gilardi, M.~Alizadeh, and M.~Kubli.
\newblock Chatgpt outperforms crowd workers for text-annotation tasks.
\newblock {\em Proceedings of the National Academy of Sciences}, 120(30):e2305016120, 2023.

\bibitem{Harris:2019}
M.~Harris, A.~Qi, L.~Jeagal, N.~Torabi, D.~Menzies, A.~Korobitsyn, M.~Pai, R.~R. Nathavitharana, and F.~Ahmad~Khan.
\newblock A systematic review of the diagnostic accuracy of artificial intelligence-based computer programs to analyze chest x-rays for pulmonary tuberculosis.
\newblock {\em PloS one}, 14(9):e0221339, 2019.

\bibitem{hu2024large}
S.~Hu, G.~Zou, S.~Yang, B.~Zhang, and Y.~Chen.
\newblock Large language model meets graph neural network in knowledge distillation.
\newblock {\em arXiv preprint arXiv:2402.05894}, 2024.

\bibitem{hyland2023maira}
S.~L. Hyland, S.~Bannur, K.~Bouzid, D.~C. Castro, M.~Ranjit, A.~Schwaighofer, F.~P{\'e}rez-Garc{\'\i}a, V.~Salvatelli, S.~Srivastav, A.~Thieme, et~al.
\newblock Maira-1: A specialised large multimodal model for radiology report generation.
\newblock {\em arXiv preprint arXiv:2311.13668}, 2023.

\bibitem{jain2021radgraph}
S.~Jain, A.~Agrawal, A.~Saporta, S.~Truong, T.~Bui, P.~Chambon, Y.~Zhang, M.~P. Lungren, A.~Y. Ng, C.~Langlotz, et~al.
\newblock Radgraph: Extracting clinical entities and relations from radiology reports.
\newblock In {\em Thirty-fifth Conference on Neural Information Processing Systems Datasets and Benchmarks Track (Round 1)}, 2021.

\bibitem{jiang2023mistral}
A.~Q. Jiang, A.~Sablayrolles, A.~Mensch, C.~Bamford, D.~S. Chaplot, D.~d.~l. Casas, F.~Bressand, G.~Lengyel, G.~Lample, L.~Saulnier, et~al.
\newblock Mistral 7b.
\newblock {\em arXiv preprint arXiv:2310.06825}, 2023.

\bibitem{jiang2024mixtral}
A.~Q. Jiang, A.~Sablayrolles, A.~Roux, A.~Mensch, B.~Savary, C.~Bamford, D.~S. Chaplot, D.~d.~l. Casas, E.~B. Hanna, F.~Bressand, et~al.
\newblock Mixtral of experts.
\newblock {\em arXiv preprint arXiv:2401.04088}, 2024.

\bibitem{johnson2019mimic}
A.~E. Johnson, T.~J. Pollard, S.~J. Berkowitz, N.~R. Greenbaum, M.~P. Lungren, C.-y. Deng, R.~G. Mark, and S.~Horng.
\newblock Mimic-cxr, a de-identified publicly available database of chest radiographs with free-text reports.
\newblock {\em Scientific data}, 6(1):317, 2019.

\bibitem{khanna2023radgraph2}
S.~Khanna, A.~Dejl, K.~Yoon, S.~Q. Truong, H.~Duong, A.~Saenz, and P.~Rajpurkar.
\newblock Radgraph2: Modeling disease progression in radiology reports via hierarchical information extraction.
\newblock In {\em Machine Learning for Healthcare Conference}, pages 381--402. PMLR, 2023.

\bibitem{lin2004rouge}
C.-Y. Lin.
\newblock Rouge: A package for automatic evaluation of summaries.
\newblock In {\em Text summarization branches out}, pages 74--81, 2004.

\bibitem{liu2023g}
Y.~Liu, D.~Iter, Y.~Xu, S.~Wang, R.~Xu, and C.~Zhu.
\newblock G-eval: Nlg evaluation using gpt-4 with better human alignment.
\newblock In {\em The 2023 Conference on Empirical Methods in Natural Language Processing}, 2023.

\bibitem{papineni2002bleu}
K.~Papineni, S.~Roukos, T.~Ward, and W.-J. Zhu.
\newblock Bleu: a method for automatic evaluation of machine translation.
\newblock In {\em Proceedings of the 40th annual meeting of the Association for Computational Linguistics}, pages 311--318, 2002.

\bibitem{smit2020combining}
A.~Smit, S.~Jain, P.~Rajpurkar, A.~Pareek, A.~Y. Ng, and M.~Lungren.
\newblock Combining automatic labelers and expert annotations for accurate radiology report labeling using bert.
\newblock In {\em Proceedings of the 2020 Conference on Empirical Methods in Natural Language Processing (EMNLP)}, pages 1500--1519, 2020.

\bibitem{team2023gemini}
G.~Team, R.~Anil, S.~Borgeaud, Y.~Wu, J.-B. Alayrac, J.~Yu, R.~Soricut, J.~Schalkwyk, A.~M. Dai, A.~Hauth, et~al.
\newblock Gemini: a family of highly capable multimodal models.
\newblock {\em arXiv preprint arXiv:2312.11805}, 2023.

\bibitem{tornberg2023chatgpt}
P.~T{\"o}rnberg.
\newblock Chatgpt-4 outperforms experts and crowd workers in annotating political twitter messages with zero-shot learning.
\newblock {\em arXiv preprint arXiv:2304.06588}, 2023.

\bibitem{touvron2023llama}
H.~Touvron, L.~Martin, K.~Stone, P.~Albert, A.~Almahairi, Y.~Babaei, N.~Bashlykov, S.~Batra, P.~Bhargava, S.~Bhosale, et~al.
\newblock Llama 2: Open foundation and fine-tuned chat models.
\newblock {\em arXiv preprint arXiv:2307.09288}, 2023.

\bibitem{wang2023chatcad}
S.~Wang, Z.~Zhao, X.~Ouyang, Q.~Wang, and D.~Shen.
\newblock Chatcad: Interactive computer-aided diagnosis on medical image using large language models.
\newblock {\em arXiv preprint arXiv:2302.07257}, 2023.

\bibitem{wei2022chain}
J.~Wei, X.~Wang, D.~Schuurmans, M.~Bosma, F.~Xia, E.~Chi, Q.~V. Le, D.~Zhou, et~al.
\newblock Chain-of-thought prompting elicits reasoning in large language models.
\newblock {\em Advances in neural information processing systems}, 35:24824--24837, 2022.

\bibitem{yu2023evaluating}
F.~Yu, M.~Endo, R.~Krishnan, I.~Pan, A.~Tsai, E.~P. Reis, E.~K. U.~N. Fonseca, H.~M.~H. Lee, Z.~S.~H. Abad, A.~Y. Ng, et~al.
\newblock Evaluating progress in automatic chest x-ray radiology report generation.
\newblock {\em Patterns}, 4(9), 2023.

\bibitem{zhang2019bertscore}
T.~Zhang, V.~Kishore, F.~Wu, K.~Q. Weinberger, and Y.~Artzi.
\newblock Bertscore: Evaluating text generation with bert.
\newblock {\em arXiv preprint arXiv:1904.09675}, 2019.

\bibitem{zheng2024judging}
L.~Zheng, W.-L. Chiang, Y.~Sheng, S.~Zhuang, Z.~Wu, Y.~Zhuang, Z.~Lin, Z.~Li, D.~Li, E.~Xing, et~al.
\newblock Judging llm-as-a-judge with mt-bench and chatbot arena.
\newblock {\em Advances in Neural Information Processing Systems}, 36, 2024.

\bibitem{zhou2023universalner}
W.~Zhou, S.~Zhang, Y.~Gu, M.~Chen, and H.~Poon.
\newblock Universalner: Targeted distillation from large language models for open named entity recognition.
\newblock {\em arXiv preprint arXiv:2308.03279}, 2023.

\end{thebibliography}
\bibliographystyle{abbrv}

\end{document}